\documentclass[10pt,twocolumn,letterpaper]{article}

\usepackage{arxiv}
\usepackage{cite}
\usepackage{bm}
\usepackage{graphicx}
\usepackage{amsmath}
\usepackage{multirow}
\usepackage{caption}
\usepackage{subfigure}
\usepackage{color}
\usepackage{times}
\usepackage{amssymb}
\usepackage{multirow}
\usepackage{subfloat}
\usepackage[breaklinks=true,bookmarks=false]{hyperref}
\cvprfinalcopy 

\usepackage[justification=centering]{caption}
\begin{document}

	\title{Prior-Knowledge and Attention based Meta-Learning for Few-Shot Learning}

	\author{Yunxiao Qin$^{1,2}$, Weiguo Zhang$^{1}$, Chenxu Zhao$^{2}$, Zezheng Wang$^{2}$, Xiangyu Zhu$^{3}$ \\ Guojun Qi$^{4}$, Jingping Shi$^{1}$, Zhen Lei$^{3}$\\
	$^1$Northwestern Polytechnical University of China $^2$AIBEE,\\
	$^3$Institute of Automation, Chinese Academy of Science, $^4$Huawei Cloud \\
	 \{qyxqyx, zhangwg, shijingping\}@mail.nwpu.edu.cn, guojunq@gmail.com, \\
	 \{xiangyu.zhu, zlei\}@nlpr.ia.ac.cn, \{cxzhao, zezhengwang\}@jd.com
}

\maketitle
	
	\begin{abstract}
Recently, meta-learning has been shown as a promising way to solve few-shot learning.
In this paper, inspired by the human cognition process which utilizes both prior-knowledge and vision attention in learning new knowledge, we present a novel paradigm of meta-learning approach with three developments to introduce attention mechanism and prior-knowledge for meta-learning.
In our approach, prior-knowledge is responsible for helping meta-learner expressing the input data into high-level representation space, and attention mechanism enables meta-learner focusing on key features of the data in the representation space.
Compared with existing meta-learning approaches which pay little attention to prior-knowledge and vision attention, our approach alleviates the meta-learner's few-shot cognition burden.
Furthermore, a Task-Over-Fitting (TOF) problem\footnotemark[1], which indicates that the meta-learner has poor generalization on different $K$-shot learning tasks, is discovered and we propose a Cross Entropy across Tasks (CET) metric\footnotemark[2] to model and solve the TOF problem.
Extensive experiments demonstrate that we improve the meta-learner with state-of-the-art performance on several few-shot learning benchmarks, and at the same time the TOF problem can also be released greatly.
		
	\end{abstract}
	
	\footnotetext[1]{When is tested on $J$-shot classification tasks, the meta-learner trained on $K$-shot tasks performs not as well as the one trained on $J$-shot tasks, where $K$ and $J$ are different unsigned integers denoting different numbers of shots for the meta-learner.}
	\footnotetext[2]{A metric for quantizing how much a meta-learning method suffers from the TOF problem.}
	
	
	\section{Introduction}
	The development of deep learning makes remarkable progresses in many tasks\cite{image-classification1, image-classification2, image-classification3, machine-trans1}.
	To achieve all of them, large amounts of thousands and even millions of labeled data are required for the deep learning approach to obtain satisfactory performance. 
	However, collecting and annotating abundant data is notoriously expensive. 
	Therefore, few-shot learning\cite{match-network,prototypical,li2019revisiting} which requires the model to learn from a few data, has attracted researchers' attention in recent years.

\begin{figure}
		\centering
		\includegraphics[width=0.9\columnwidth]{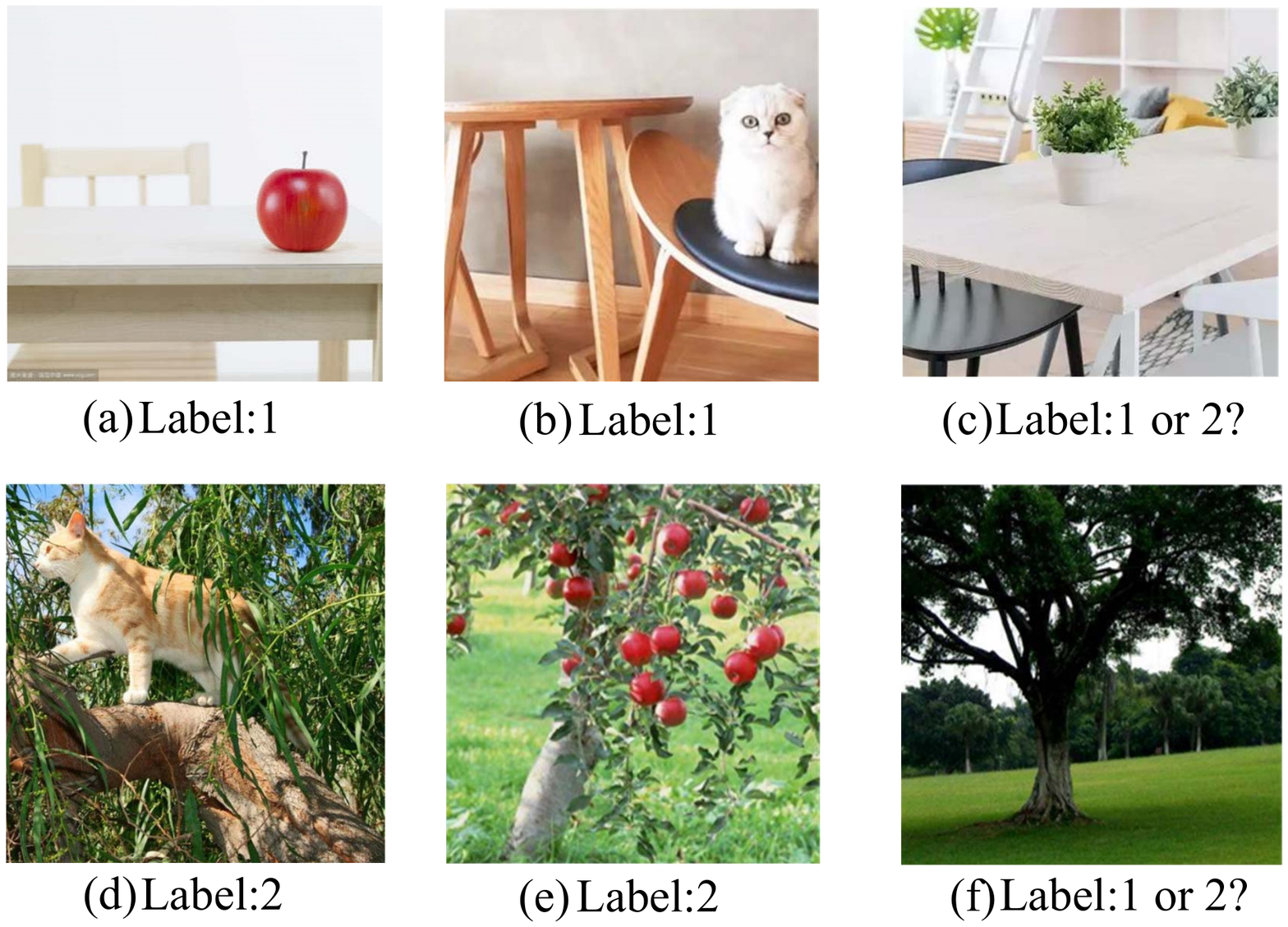}
		\caption{
			An Example of few-shot classification task. 
			The six images come from two classes, where four labeled ones are training data with the two unlabeled for test. 
			When predicting the two testing images, we utilize our prior-knowledge about the world to understand all components in these images and use our vision attention to pay attention to key components table and tree.
			Finally, we predict the image (c) belongs to class 1 that contains table, while the image (f) is associated with class 2 of tree. 
		}
		\label{fig:example to understand idea}
	\end{figure}

	Learning from few-data is challenging for Computer Vision. 
	In comparison, we human beings can rapidly learn new categories from very few examples.
	Recently, meta-learning\cite{Learning-a-synaptic, On-the-optimiazation, An-alternative-to, MAML, Meta-SGD, miniimagenet, Reptile, SNAIL, zhuang, RL2, Metagan, metanet} has shown promising performance to improve the few-shot learning for Computer Vision. 
	However, existing meta-learning methods commonly ignore prior-knowledge\cite{langer1981prior,dochy1990instructional,shapiro2004including,wylie2004interactive,hsin2015effects} and attention mechanism\cite{human_attention1,human_attention2} which have been both demonstrated important for human cognitive and learning process.
	We illustrate a few-shot classification problem in Fig.\ref{fig:example to understand idea} for a better understanding of the role of prior-knowledge and attention mechanism in human few-shot learning. 
	In Fig.\ref{fig:example to understand idea}, we unconsciously leverage our learned knowledge about the world to understand and express these images into high-level compact representations, such as plant, animal, tree, and table \emph{etc.} 
	However, according to the four training images, we discover that only the feature of the tree and table are useful for us to recognize these two classes of images. 
	Then, we quickly adjust ourselves to pay attention to the critical features and make the decision based on the focused features. 
	
	Evidently, we can summarize two main modules in human few-shot learning: \textbf{a stable Representation module that utilizes prior-knowledge to express the image into compact feature representations; and a smart attention-based decision logical module that adapts accurately and performs recognition based on the feature representations}.
	While existing meta-learning approaches commonly train meta-learners to learn adaptive networks directly based on the original input data with no attention mechanism and prior-knowledge.
	
	In this paper, inspired by the human cognition process, we present a novel paradigm of meta-learning approach with three developments to introduce attention mechanism and prior-knowledge step-by-step for meta-learning.
	Here, we briefly introduce the proposed methods.
	\textbf{1)} The first method is \textbf{A}ttention based \textbf{M}eta-\textbf{L}earning (AML) which leverages attention mechanism to enable the meta-learner paying more attention on essential feature.
	\textbf{2)} For the meta-learner enjoying not only attention but also prior-knowledge, we present another method \textbf{R}epresentation and \textbf{A}ttention based \textbf{M}eta-\textbf{L}earning (RAML).
	Its network contains a Representation module and an attention-based prediction (ABP) module. 
	The Representation module is similar to the same module of human vision.
	It learns the prior-knowledge in a supervised fashion and is responsible for understanding and extracting stable compact feature representations from the input image. 
	The ABP module plays the same role as the smart attention-based decision logic module of human vision.
	It enables the meta-learner to precisely adjusting first its attention to the most discriminative feature representations of input images and second the corresponding predictions.
	\textbf{3)} In the third method, to take full advantage of endless unlabeled data, we design a novel method where the Representation module learns the past knowledge in unsupervised fashion \cite{AE,VAE,context-encoder, colorization,deep-cluster,split-brain}. 
	We call this method \textbf{U}nsupervised \textbf{R}epresentation and \textbf{A}ttention based \textbf{M}eta-\textbf{L}earning(URAML). 
	With URAML, we show in our experiments that the growth of the number of unlabeled data and the development of unsupervised learning both improve the performance of URAML apparently. 
	
	In addition, we show a Task-Over-Fitting (TOF) problem for existing meta-learning and present a Cross-Entropy across Tasks (CET) metric to evaluate how much a meta-learning method is troubled by the TOF problem.
	An example of the TOF problem is, the meta-learner trained on 5-way 1-shot tasks is not as capable as the one trained on 5-way 5-shot tasks when they are tested on 5-way 5-shot tasks, and vice versa.
	However, in practical applications, it is uncertain how much data and how many shot times are available to the meta-learner to learn. 
	Therefore, we argue that the trained meta-learner should generalizes well to different $K$-shot tasks.
	The possible reason behind the TOF problem is that existing meta-learners are vulnerable to the features irrelevant to the presented tasks since they ignoring both priori knowledge and attention mechanism.
	Our experiment validates that by incorporating prior-knowledge and attention mechanism, our methods suffer less from the TOF problem than existing meta-learning methods.
	
	We summarize the main contributions of our work as:
	\begin{itemize}
		\item We propose that both attention mechanism and prior-knowledge are crucial for meta-learner to reduce its cognition burden in few-shot learning, and we develop three methods AML, RAML, and URAML to step-by-step leverage attention mechanism and prior-knowledge in meta-learning.
		
		\item We discover the TOF problem for meta-learning, and design a novel metric Cross-Entropy across Tasks (CET) to measure how much meta-learning approaches suffer from the TOF problem.
		
		\item Through extensive experiments, we show that the proposed methods achieve state-of-the-art performance on several few-shot learning benchmarks and in the meantime, they are less sensitive to the TOF problem, especially the RAML and URAML.
		
	\end{itemize}

	\section{Related Work}
	\subsection{Meta-learning for Few-Shot Learning}
	An $N$-way $K$-shot learning task contains a support set and a query set. 
	The support and query set contain $K$ and $L$ examples for each of the $N$ classes, respectively.
	Existing meta-learning approaches usually solve the few-shot learning by training a meta-learner on the $N$-way $K$-shot learning tasks in the following way.
	Firstly, the meta-learner is required to inner-update itself on the support set.
	Secondly, after the inner-updating, meta-learner is evaluated on the query set.
	Finally, by minimizing the loss on the query set, the meta-learner learns a base learner which has easy-fine-tune weights\cite{MAML,Reptile} or a skillful weight updater\cite{miniimagenet, metanet} or both\cite{Meta-SGD} or the ability to memorize the support set\cite{SNAIL}.
	The methods train the meta-learner learning an easy-fine-tune base learner are also called as weight initialization based methods, as the meta-learner learns generalized initial weight for few-shot learning tasks.
	Recently, MAML, which is a classical weight initialization based method, is popular and lots of MAML based methods have been proposed.
	For example, LLAML\cite{LLAML} uses a local Laplace approximation to model the task parameters, and MTL\cite{sun2019meta-transfer} trains a meta-transfer to adapt a pre-trained deep network to few-shot learning tasks.
	Besides, MetaGAN\cite{Metagan} shows that by coupling MAML with adversarial training, the meta-learner is trained to learn a better decision boundaries between different classes in few-shot learning.
	To reduce the computation and memory cost of MAML, iMAML\cite{IMAML} leverages implicit differentiation to remove the need of differentiation through the inner-update path.
	
	Though existing meta-learning methods performs promising, they seldom consider the prior-knowledge and attention mechanism in meta-learning.
	In our paper, we improve meta-learning for few-shot learning by introducing prior-knowledge and attention mechanism to meta-learning.
	
	\subsection{Attention Mechanism}
	Recent years, attention mechanism\cite{soft_attention1,soft_attention3,hard_attention,attention-is-all-you-need} has been widely used in computer vision systems, machine translation and \emph{etc.}. 
	Several manners of the attention mechanism have been proposed, such as soft attention\cite{soft_attention1,soft_attention3}, hard attention\cite{hard_attention} and self attention\cite{attention-is-all-you-need} \emph{etc.} 
	Soft attention can be seen as simulating the attention mechanism by multiplying weight on the neural unit so that the network pays more attention on the neural unit which multiplies with larger weight.
	SENet\cite{soft_attention3} takes advantage of soft attention mechanism to win the champion on the image classification task of ILSVRC-2017\cite{berg2010large}. 
	Hard attention\cite{hard_attention} can be seen as a module that decides a block region of the input image where is visible to the network, and the other region is invisible. 
	Self-attention\cite{attention-is-all-you-need} improves the performance of the machine translation system by training a network to find the inner dependency of the input and that of the output. 
	In this paper, we use soft attention mechanism as the meta-learner's attention mechanism.

	\subsection{Unsupervised Representation Learning}
	Supervised learning is a data-hungry manner to train deep network. 
	Considering this, several unsupervised learning approach\cite{AE,VAE,context-encoder,colorization,deep-cluster,split-brain} have been proposed. 
	A well-known way is training a neural network to reconstruct the original input through an Encoder-Decoder architecture, such as Auto-Encoder\cite{AE}, Variational Auto-Encoder (VAE)\cite{VAE} and \emph{etc}. 
	Given partial masked images, Context Auto-Encoder\cite{context-encoder} trains a network to reconstruct not only the visible but also the masked region of the image.
	Colorization\cite{colorization} uses \emph{Lab} images to train a network to generate the unseen \emph{ab} channels from the input \emph{L} channel. 
	Based on Colorization, Split-Brain\cite{split-brain} trains two separated networks to separately generate the \emph{ab} channels from the \emph{L} channel and generate the \emph{L} channel from the \emph{ab} channels. 
	Different from these methods, DeepCluster\cite{deep-cluster} couples deep learning with Cluster algorithm\cite{cluster1,cluster2}.
	However, in real world, many unlabeled images containing complex semantic information and are not suitable to be categorized into specific clusters.
	Therefore, we consider there might be a limitation for DeepCluster and we utilize Split-Brain as the unsupervised learning method in URAML.
	
	\section{Method}
	\subsection{Problem of Learning from Few-Data}
	Learning from few-data is extremely difficult for the deep learning model. 
	One reason is that the original input data is commonly represented in a large dimension space. 
	Usually, tens or hundreds of thousands of dimension space is required. 
	For example, for the image classification task, the original image is commonly stored in a large dimensional space (dimension of an 224x224 RGB image is 150528).
	In such a large dimension space, it is difficult for a few samples of one category to accurately reflect the character of this category. 
	
	Humans learn new categories efficiently because they utilize prior-knowledge and attention mechanism in cognition\cite{langer1981prior,dochy1990instructional,ormerod1990human,oliva2003top,van2003selective,posner1990attention,ungerleider2000mechanisms,hsin2015effects,wylie2004interactive}. 
	Prior-knowledge facilitates human to express perceptual images into high-level representations or descriptions, and attention mechanism helps human to focus on critical components of the representations.
	In this way, humans reduce the dimension of images and maintain the discriminative components of the images, which alleviates human cognition load and facilitate humans to efficiently learn new categories.
	
	Existing meta-learning approaches improve deep learning a lot in few-shot learning. 
	However, they train the meta-learner to quickly fit few-shot learning tasks directly on the few original high dimensional input data and pay little attention to the importance of prior-knowledge and attention mechanism, leading to unsatisfactory performance. 
	Besides, as introduced before, we propose that ignoring prior-knowledge and attention mechanism is also the possible reason for existing meta-learning approaches to be vulnerable to suffer from the TOF problem.
	
	In this paper, inspired by human cognition and for addressing the problem existing meta-learning approaches expose, we propose three methods step-by-step: Attention based Meta-Learning (AML), Representation and Attention based Meta-Learning (RAML), Unsupervised Representation and Attention based Meta-Learning (URAML).

	\subsection{AML}
	AML equips the meta-learner with the power of attention mechanism.
	We first introduce the network structure and then detail the training of AML.
	
	\noindent \textbf{AML Network} \quad
	
\begin{figure}
		\centering
		\includegraphics[width=0.9\columnwidth]{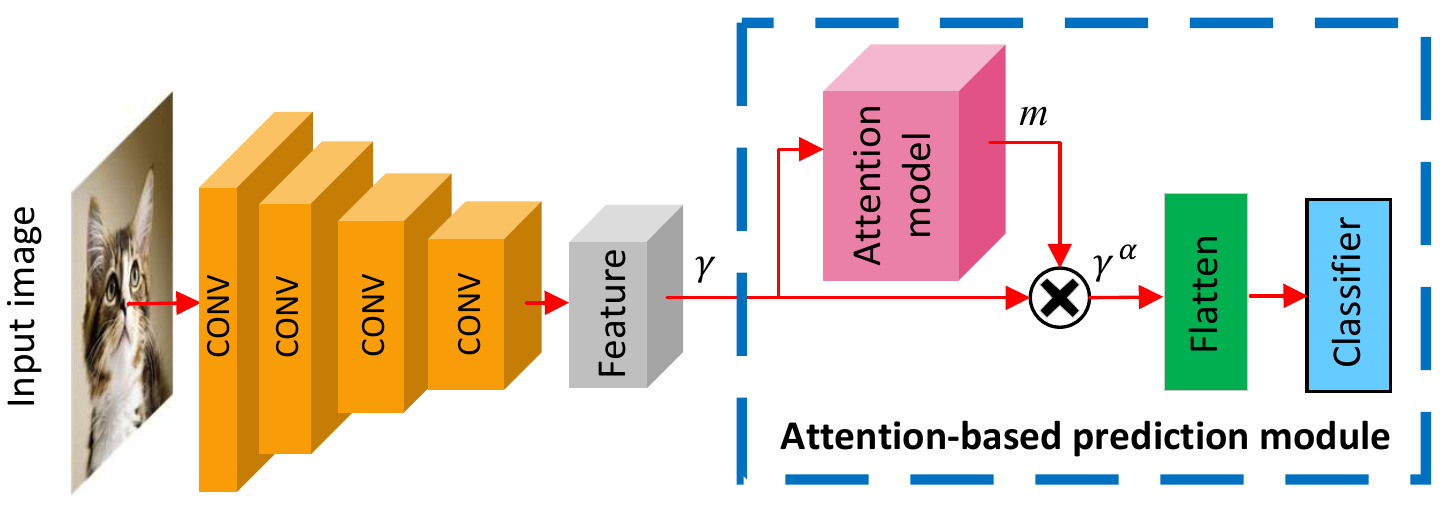}
		\caption{
			Network structure of the proposed method AML. 
			There is an attention model inserted explicitly in the meta-learner's network.}
		\label{fig:network of the AML method}
	\end{figure}

	The network architecture of AML is shown in Fig.\ref{fig:network of the AML method}.
	The network consists of a feature extractor and an attention-based prediction (ABP) module.
	The feature extractor is a CNN $\mathcal{F}$ which is composed of four stacking convolutional layers.
	The ABP module contains an convolution-based attention model $\mathcal{A}$ and a fully-connect layer based classifier $\mathcal{C}$.
	Eq.\ref{eq:AML} shows the inference of the network.
	$\theta_{f}$, $\theta_{a}$, and $\theta_{c}$ are weights of $\mathcal{F}$, $\mathcal{A}$, and $\mathcal{C}$, respectively. 
	$\mathcal{F}$ extracts features $\gamma_i$ of the input image $x_i$ and feed $\gamma_i$ into the attention model $\mathcal{A}$.
	Then, $\mathcal{A}$ calculates the soft attention mask $m_i$ of the features $\gamma_i$.
	By channel-wise multiplication $\odot$ between $\gamma_i$ and $m_i$, the focused features $\gamma^\alpha_i$ is calculated.
	Finally, the classifier $\mathcal{C}$ predicts the category of the input image, and $\hat{y_i}$ is the corresponding prediction of $x_i$.
	We simplify and integrate the inference in Eq.\ref{eq:AML} as $\hat{y_i} = \mathbb{F}(x_i; \theta_f, \theta_a, \theta_c)$.
	
	\begin{equation}
	\left\{
	\begin{array}{lr}
	\gamma_i = \mathcal{F}(x_i; \ \theta_{f}) \\
	m_i = \mathcal{A}(\gamma_i; \ \theta_{a}) \\
	\gamma^\alpha_i = \gamma_i \odot m_i \\
	\hat{y_i} = \mathcal{C}(\gamma^\alpha_i; \ \theta_{c})
	\end{array}
	\right.
	\label{eq:AML}
	\end{equation}
	
	In this paper, we use soft attention mechanism to build up the attention model.
	Although the soft attention mechanism is not exactly the same with the attention mechanism in human vision, it still plays a similar role with the human attention mechanism and helps the meta-learner to control its attention to key features.
	Fig.\ref{fig:4b} is used to better understand the soft attention processing of the meta-learner. 
	
\begin{figure}
		\centering
		\includegraphics[width=0.7\columnwidth]{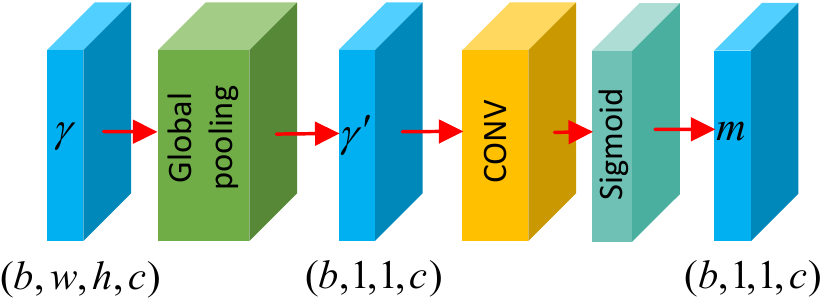}
		\caption{
			Inner network structure of the attention model. 
			The shape of feature map $\gamma$ is (\emph{b,w,h,c}) which is shown at the left of the figure, where \emph{b, w, h, c} are the batch size, width, height and umber of channels of the feature map $\gamma$, and the shape of $\gamma'$ and \emph{m} are both (\emph{b},1,1,\emph{c}).
		}
		\label{fig:network structure of attention}
	\end{figure}

\begin{figure*}[t]
		\centering    
		\subfigure[] 
		{
			\includegraphics[scale=0.66]{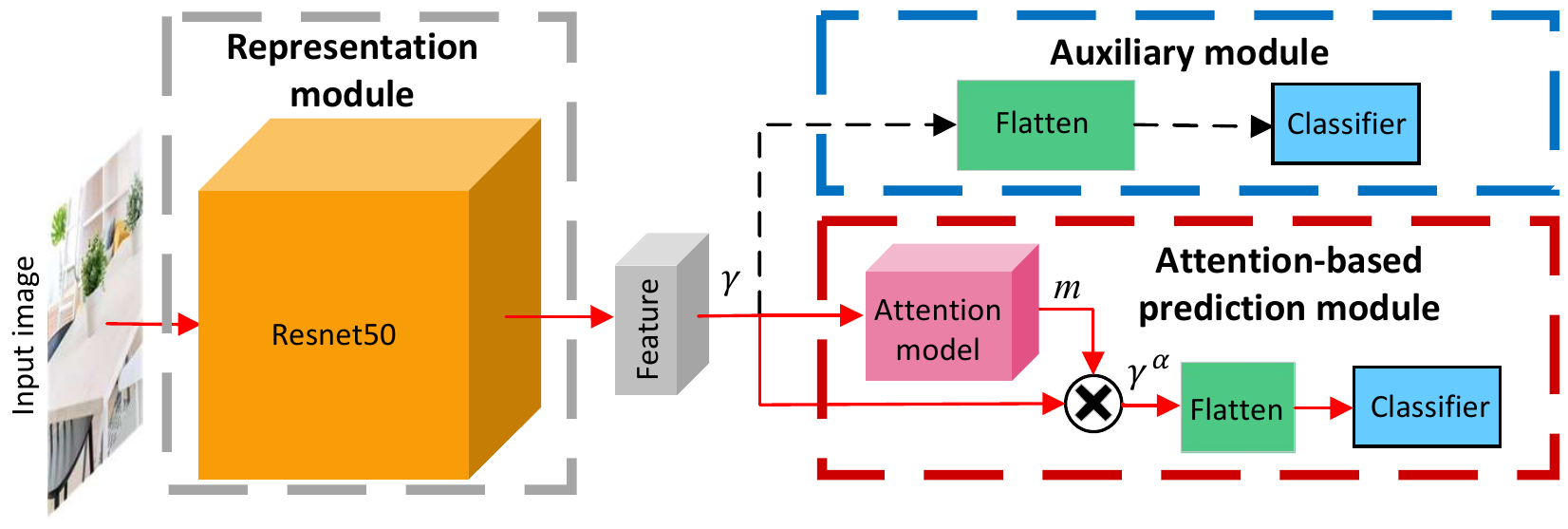}  
			\label{fig:4a} 
		}
		\subfigure[] 
		{
			\includegraphics[scale=0.66]{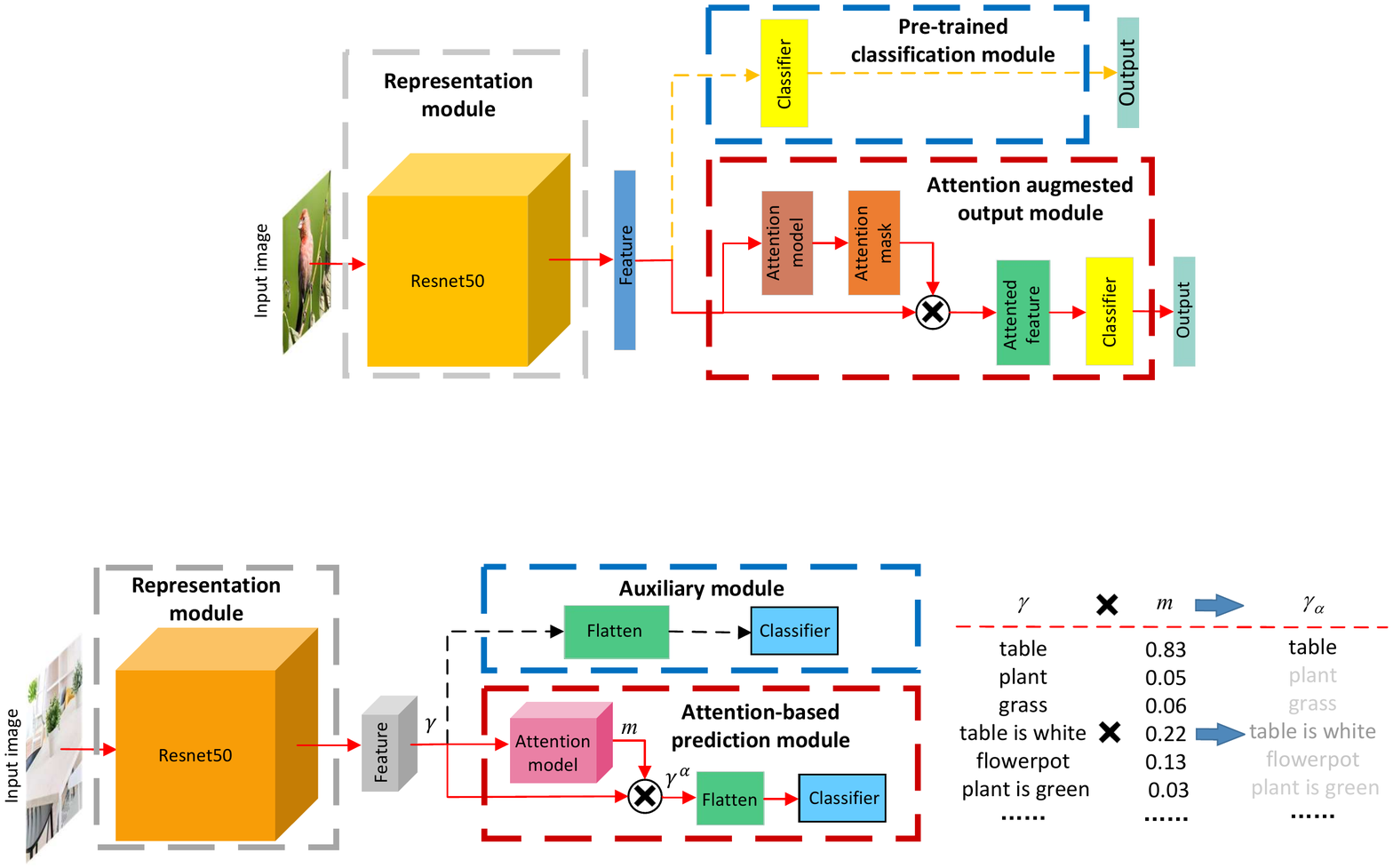}
			\label{fig:4b}
		}
		\caption{
			(a) Network structure of the proposed RAML. 
			The meta-learner is composed of a Representation module and an ABP module. 
			The Auxiliary module is used to assist the meta-learner to learn prior-knowledge.
			(b) An example that interprets the principle of soft attention mechanism for few-shot learning.} 
		\label{fig:network of the RAML method}  
	\end{figure*}

	Fig.\ref{fig:network structure of attention} shows the attention model structure and Eq.\ref{eq:attention model} shows the inference of the attention model. 
	The input feature $\gamma$ is firstly global-average-pooled to get feature $\gamma$$'$, and then a convolution layer coupled with a sigmoid activation layer are used to predict the attention mask \emph{m} from the feature $\gamma$$'$. 
	
	\begin{equation}
	\left\{
	\begin{array}{lr}
	\gamma' = \mathcal{P}_{a}(\gamma), \\
	m = \sigma(\mathcal{F}_{a}(\gamma'; \ \theta_{a})) 
	\end{array}
	\right.
	\label{eq:attention model}
	\end{equation}
	
	$\mathcal{P}_{a}$ is the global-average-pooling operation, and $\sigma$ is the sigmoid activation, and $\mathcal{F}_{a}$ is the convolution layer in the attention model.

	\noindent \textbf{AML Meta-Train Process} \quad
	
	Given a few-shot classification task $\tau$, AML meta-trains the meta-learner to solve the task $\tau$ in the two steps.
	\textbf{First}, AML requires the meta-learner to inner-update itself on the the support set of $\tau$, which can be formulated as Eq.\ref{eq:AML_support_update1} and Eq.\ref{eq:AML_support_update2}.
	\begin{equation}
	\left\{
	\begin{array}{lr}
	\hat{y_i} = \mathbb{F}(x_i; \theta_f, \theta_a, \theta_c), \\
	\mathcal{L}_i(\theta_f, \theta_a, \theta_c) = l(\hat{y_i}, y_i), \\
	\mathfrak{L}_s(\theta_f, \theta_a, \theta_c) = \frac{1}{N_s} \displaystyle{\sum_{i=1}^{N_s}} \mathcal{L}_i(\theta_f, \theta_a, \theta_c) 
	\end{array}
	\right.
	\label{eq:AML_support_update1}
	\end{equation}
	\begin{equation}
	(\theta^{'}_f, \theta^{'}_a, \theta^{'}_c) = (\theta_f, \theta_a, \theta_c) - \alpha{\boldmath \circ}\nabla_{(\theta_f, \theta_a, \theta_c)}\mathfrak{L}_{s}(\theta_f, \theta_a, \theta_c)
	\label{eq:AML_support_update2}
	\end{equation}
	
	In Eq.\ref{eq:AML_support_update1}, $x_i$ is any image that belongs to the support set, $l$ is the cross-entropy loss function, $\mathcal{L}_i$ is the meta-learner's loss on the image $x_i$, $\mathfrak{L}_s$ is the meta-learner's loss on the total support set, and $N_s$ is the number of images in the support set.
	In Eq.\ref{eq:AML_support_update2}, inspired by Meta-SGD\cite{Meta-SGD}, we set $\alpha$ as a trainable vector which adjusts the inner-update direction and $\alpha$ has the same shape with the weights $\theta_f, \theta_a$, and $\theta_c$.
	$\alpha$ can also be presented as $\alpha = [\alpha_f, \alpha_a, \alpha_c]$ and the Eq.\ref{eq:AML_support_update2} can be split into three equations, \emph{i.e.} $\theta^{'}_f = \theta_f - \alpha_f{\boldmath \circ}\nabla_{\theta_f}\mathfrak{L}_{s}(\theta_f, \theta_a, \theta_c)$ and \emph{etc.}.
	For simplicity, we merge these three equations into one equation as Eq.\ref{eq:AML_support_update2} shows.
	$\circ$ is the element-wise multiplication.
	Supervised by the loss on the support set, the meta-learner inner-updates its weights $\theta_f, \theta_a, \theta_c$ to $\theta^{'}_f, \theta^{'}_a, \theta^{'}_c$.
	
	\textbf{Second}, as the inner-updated weight $\theta^{'}_f, \theta^{'}_a$, and $\theta^{'}_c$ depend on not only the initial values of $\theta_f, \theta_a$, and $\theta_c$, but also $\alpha$, all 
	$\theta_f, \theta_a, \theta_c$, and $\alpha$ can be meta-optimized.
	We formulate this process as Eq.\ref{eq:AML_support_optimize1} and Eq.\ref{eq:AML_support_optimize2}.
	\begin{equation}
	\left\{
	\begin{array}{lr}
	\hat{y_i} = \mathbb{F}(x_i; \theta^{'}_f, \theta^{'}_a, \theta^{'}_c), \\
	\mathcal{L}_i(\theta^{'}_f, \theta^{'}_a, \theta^{'}_c) = l(\hat{y_i}, y_i), \\
	\mathfrak{L}_q(\theta^{'}_f, \theta^{'}_a, \theta^{'}_c) = \frac{1}{N_q} \displaystyle{\sum_{i=1}^{N_q}} \mathcal{L}_i(\theta^{'}_f, \theta^{'}_a, \theta^{'}_c) 
	\end{array}
	\right.
	\label{eq:AML_support_optimize1}
	\end{equation}
	\begin{equation}
	(\theta_f, \theta_a, \theta_c, \alpha) = (\theta_f, \theta_a, \theta_c, \alpha) - \beta{\boldmath \cdot}\nabla_{(\theta_f, \theta_a, \theta_c, \alpha)}\mathfrak{L}_{q}(\theta^{'}_f, \theta^{'}_a, \theta^{'}_c)
	\label{eq:AML_support_optimize2}
	\end{equation}
	
	In Eq.\ref{eq:AML_support_optimize1}, $x_i$ is an image belonging to the query set, and $N_q$ denotes the number of images in the query set.
	$\mathfrak{L}_q$ is the inner-updated meta-learner's loss on the query set.
	It should be noted that $\nabla_{(\theta_f, \theta_a, \theta_c, \alpha)}\mathfrak{L}_{q}(\theta^{'}_f, \theta^{'}_a, \theta^{'}_c)$ computes the gradient of $\mathfrak{L}_{q}$ towards $(\theta_f, \theta_a, \theta_c, \alpha)$ but not $(\theta^{'}_f, \theta^{'}_a, \theta^{'}_c)$.
	By optimizing $\mathfrak{L}_q$, the meta-learner is forced to learn not only the suitable initial weights $\theta_f, \theta_a, \theta_c$ but also $\alpha$ for task $\tau$.
	With the learned initial weights and $\alpha$, the meta-learner can inner-update itself precisely on the support set and then perform well on the query set.

	In AML, the meta-learner is trained on lots of few-shot learning tasks with these two steps, which makes the meta-learner learn generalizable initial weights for not only the feature extractor $\mathcal{F}$ and the classifier $\mathcal{C}$, but also the attention model $\mathcal{A}$.
	While existing initialization based meta-learning methods only train the meta-learner to learn initial weights for the feature extractor and the classifier.
	Therefore, compared with existing meta-learners, AML simplifies the few-shot problem and improves performance since its attention ability is meta-trained and can be easily adjusted to the crucial features for solving few-shot learning, which leads the classifier can make a precise prediction for the input.
	In our experiment, we show the positive effect of attention mechanism.

%
	
	\subsection{RAML}
	
	RAML assembles the meta-learner not only the attention mechanism but also the ability to well use the past learned knowledge. 
	
	Fig.\ref{fig:4a} shows the meta-learner's network structure.
	Its network consists of a Representation module and an ABP module.
	The Representation module is different from the feature extractor in AML because the Representation module here is responsible for the meta-learner learning and leveraging prior-knowledge to understand the input image.
	While the feature extractor in AML is meta-trained for learning how to update itself for solving few-shot learning tasks.
	In our work, the Representation module is a ResNet-50 network.
	Similar to the ABP module in AML, the ABP module here also contains an attention model and a classifier.
	It is responsible for quickly adjusting the meta-learner's attention and prediction based on the output feature from the Representation module.
	Besides, Fig.\ref{fig:4a} contains an Auxiliary module.
	The Auxiliary module does not belong to the meta-learner, and it is only used to assist the meta-learner learning prior-knowledge. 
	
	\noindent \textbf{RAML Training Process} \quad
	
	The training process of RAML can be separated into two stages: prior-knowledge learning and meta-training stage.
	
	\textbf{At the prior-knowledge learning stage}, with the assist of the Auxiliary module, the Representation module is trained to learn prior-knowledge about image classification in a supervised manner.
	The training process can be formulated as 
	\begin{equation}
	\left\{
	\begin{array}{lr}
	\gamma_i = \mathcal{F}_r(x_i; \ \theta_r) \\
	\hat{y_i} = \mathcal{C}_{au}(\gamma_i; \ \theta_{au}) \\
	L_{au} = \frac{1}{n} \displaystyle{\sum_{i=1}^{n}} l (\hat{y_i}, y_i) \\
	\theta^*_r, \theta^{*}_{au} = \mathop{argmin}\limits_{\theta_r, \theta_{au}} L_{au} 
	\end{array}
	\right.
	\label{eq:RAML_prior_train}
	\end{equation}
	
	$\mathcal{F}_r$ and $\mathcal{C}_{au}$ denote the Representation and Auxiliary modules, respectively, and $\theta_r$ and $\theta_{au}$ are their weights.
	$x_i$ is an input image used for the representation model learning prior-knowledge, and $n$ is the number of images.
	$\theta^*_r$ and $\theta^*_{au}$ are the learned values of $\theta_r$ and $\theta_{au}$.
	
	\textbf{At the meta-training stage}, for the meta-learner well using the learned knowledge to stably express the input image into high-level representations, the Representation module will not be meta-trained. 
	Similar to AML, in RAML, we simplify the prediction of the meta-learner as $\hat{y_i} = \mathbb{F}(x_i; \theta_r^*, \theta_a, \theta_c)$, where all symbols denote the same meanings as those in AML.
	In RAML, the inner-update of the meta-learner on the support set can be formulated as Eq.\ref{eq:RAML_support_update1} and Eq.\ref{eq:RAML_support_update2}.
	We can see that different from the inner-update of AML which update all weights of the network, the inner-update of RAML only update the weights $\theta_a$ and $\theta_c$ of the ABP module.
	The weight $\theta_r^*$ of the Representation module is fixed to keep the learned prior-knowledge.
	\begin{equation}
	\left\{
	\begin{array}{lr}
	\hat{y_i} = \mathbb{F}(x_i; \theta_r^*, \theta_a, \theta_c), \\
	\mathcal{L}_i(\theta_r^*, \theta_a, \theta_c) = l(\hat{y_i}, y_i), \\
	\mathfrak{L}_s(\theta_r^*, \theta_a, \theta_c) = \frac{1}{N_s} \displaystyle{\sum_{i=1}^{N_s}} \mathcal{L}_i(\theta_r^*, \theta_a, \theta_c) 
	\end{array}
	\right.
	\label{eq:RAML_support_update1}
	\end{equation}
	\begin{equation}
	(\theta^{'}_a, \theta^{'}_c) = (\theta_a, \theta_c) - \alpha{\boldmath \circ}\nabla_{(\theta_a, \theta_c)}\mathfrak{L}_{s}(\theta_r^*, \theta_a, \theta_c)
	\label{eq:RAML_support_update2}
	\end{equation}
	
	The meta-optimizing in RAML can be formulated as Eq.\ref{eq:RAML_meta_optimize1} and Eq.\ref{eq:RAML_meta_optimize2}.
	\begin{equation}
	\left\{
	\begin{array}{lr}
	\hat{y_i} = \mathbb{F}(x_i; \theta_r^*, \theta^{'}_a, \theta^{'}_c), \\
	\mathcal{L}_i(\theta_r^*, \theta^{'}_a, \theta^{'}_c) = l(\hat{y_i}, y_i), \\
	\mathfrak{L}_q(\theta_r^*, \theta^{'}_a, \theta^{'}_c) = \frac{1}{N_q} \displaystyle{\sum_{i=1}^{N_q}} \mathcal{L}_i(\theta_r^*, \theta^{'}_a, \theta^{'}_c) 
	\end{array}
	\right.
	\label{eq:RAML_meta_optimize1}
	\end{equation}
	\begin{equation}
	(\theta_a, \theta_c, \alpha) = (\theta_a, \theta_c, \alpha) - \beta{\boldmath \cdot}\nabla_{(\theta_a, \theta_c, \alpha)}\mathfrak{L}_{q}(\theta_r^*, \theta^{'}_a, \theta^{'}_c)
	\label{eq:RAML_meta_optimize2}
	\end{equation}
	
	The character of RAML is that the Representation module and the ABP module are trained separately.
	The Representation module is supervisorily trained to learn the prior-knowledge about image classification, and the ABP module is meta-trained to learn how to adjust itself quickly to solve few-shot learning tasks in the representation space provided by the Representation module. 
	Compared with AML, which meta-trains the meta-learner not only adjusting the feature extractor but also the ABP module, RAML meta-trains the meta-learner simplify the few-shot learning problem as the meta-learner only need to adjust its ABP module in the representation space.
	This is possibly the reason why RAML outperforms AML in our experiment.

	\subsection{URAML}
	
	\begin{figure*}
		\centering
		\includegraphics[width=1.98\columnwidth]{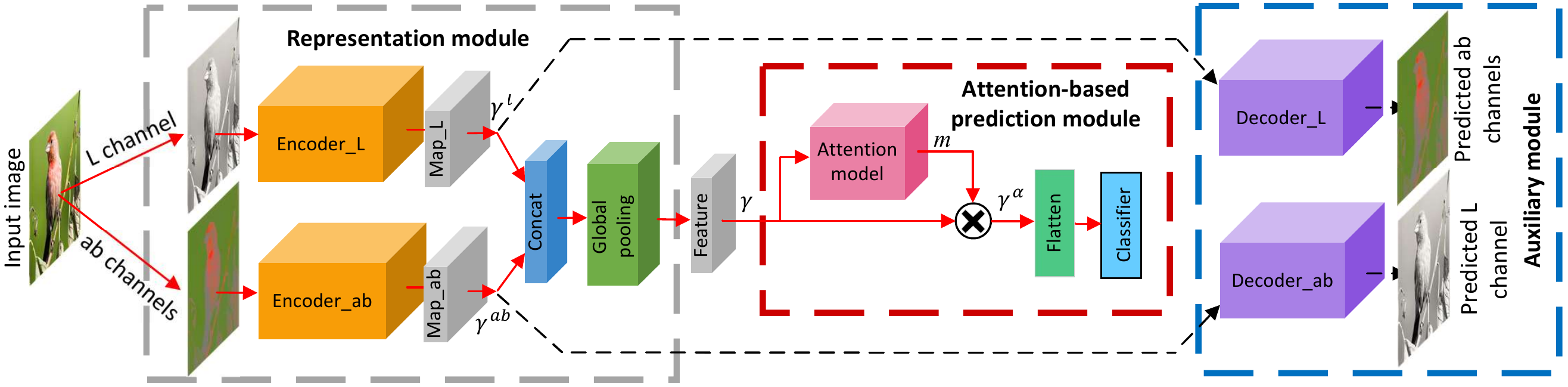}
		\caption{
			The network structure of URAML. 
			The meta-learner is composed of a Representation module and an ABP module. 
			The Auxiliary module is used to assist the meta-learner to learn prior-knowledge.
		}
		\label{fig:figure5}
	\end{figure*}
	
	The prior-knowledge can be learned on not only labeled data but also large-scale unlabelled data.
	Thus, we design the method URAML and show its network structure in Fig.\ref{fig:figure5}. 
	Similar to RAML, the meta-learner is also composed of a Representation module and an ABP module, and the Auxiliary module does not belong to the meta-learner.
	The training process of URAML can be separated into two stages: prior-knowledge learning and meta-training stage.
	
	\textbf{At the prior-knowledge learning stage}, the Representation module learns the knowledge with an unsupervised learning algorithm: Split-Brain auto-encoder\cite{split-brain}.
	The Split-Brain auto-encoder simultaneously trains two auto-encoders with \emph{Lab} images.
	In \emph{Lab} color system, the \emph{L} channel determines the brightness of the image, and the \emph{ab} channels determine the color.
	One auto-encoder in Split-Brain is trained to predict the unseen \emph{ab} channels of the input \emph{Lab} image, given only the \emph{L} channel.
	Another is trained to predict the unseen \emph{L} channel, given the \emph{ab} channels.
	As Fig.\ref{fig:figure5} shows, the Representation module consists of two ResNet-50 based encoders and the Auxiliary module consists of two corresponding deconvolution\cite{deconv} based decoders.
	We formulate the prior-knowledge learning process as Eq.\ref{eq:URAML_knowledge1} and Eq.\ref{eq:URAML_knowledge2}.
	
	\begin{equation}
	\left\{
	\begin{array}{lr}
	\gamma_i^l = \mathcal{F}_{l}(x_i^l; \ \theta_{l}) \\
	\hat{x}^{ab}_i = \mathcal{D}_l(\gamma_i^l; \ \omega_{l}) \\
	L_{l}(\theta_l, \omega_{l}) = \frac{1}{n} \displaystyle{\sum_{i=1}^{n}} l_2(x_i^{ab}, \ \hat{x}_i^{ab}) \\
	\theta_l^{*}, \omega_{l}^{*} = \mathop{argmin}\limits_{\theta_l, \omega_{l}} L_{l}(\theta_l, \omega_{l}) 
	\end{array}
	\right.
	\label{eq:URAML_knowledge1}
	\end{equation}
	
	In Eq.\ref{eq:URAML_knowledge1}, $x^l_i$ and $x^{ab}_i$ are the \emph{L} and \emph{ab} channels of the input \emph{Lab} image $x_i$, respectively.
	$\mathcal{F}_l$ and $\mathcal{D}_l$ are the encoder and decoder that predict $x^{ab}_i$ based on $x^l_i$, respectively, and $\hat{x}_i^{ab}$ is the prediction.
	$\theta_l$ and $\omega_{l}$ are the weights of $\mathcal{F}_l$ and $\mathcal{D}_l$, respectively, and $\theta_l^{*}$ and $\omega_{l}^{*}$ are the optimized values of $\theta_l$ and $\omega_{l}$.
	$\gamma_i^l$ is the squeezed feature of $x^l_i$ by the encoder $\mathcal{F}_l$.
	$L_l$ is the loss of $\mathcal{F}_l$ and $\mathcal{D}_l$, and $l_2$ is the \emph{MSE} loss function.
	$n$ is the number of \emph{Lab} images that trains $\mathcal{F}_l$ and $\mathcal{D}_l$.
	In Eq.\ref{eq:URAML_knowledge2}, all symbols are defined in the same way with those in Eq.\ref{eq:URAML_knowledge1}.
	\begin{equation}
	\left\{
	\begin{array}{lr}
	\gamma_i^{ab} = \mathcal{F}_{ab}(x_i^{ab}; \ \theta_{ab}) \\
	\hat{x}_i^l = \mathcal{D}_i^{ab}(\gamma_i^{ab}; \ \omega_{ab}) \\
	L_{ab}(\theta_{ab}, \omega_{ab}) = \frac{1}{n} \displaystyle{\sum_{i=1}^{n}} l_2(x_i^{l}, \ \hat{x}_i^{l})  \\
	\theta_{ab}^{*}, \omega_{ab}^{*} = \mathop{argmin}\limits_{\theta_{ab}, \omega_{ab}} L_{ab}(\theta_{ab}, \omega_{ab}) 
	\end{array}
	\right.
	\label{eq:URAML_knowledge2}
	\end{equation}

	After unsupervised learning, the representations $\gamma_i$ of an \emph{Lab} image $x_i$ can be calculated by first concatenating $\gamma_i^l$ with $\gamma_i^{ab}$ and second average-pooling, which is shown as $\gamma_i = \mathcal{P}_{a}(\gamma_i^l, \gamma_i^{ab})$, where $\mathcal{P}_{a}$ is an average-pooling layer.
	
	\textbf{At the meta-training stage}, the ABP module is trained in the same way with that in RAML.
	Note that, the learned weight of the Representation module in URAML is $\theta_r^* = [\theta_{l}^*, \theta_{ab}^*]$.
	
	At the end of our methodology, we summarize our three methods briefly.
	Inspired by human cognition which makes full use of attention mechanism and prior-knowledge to efficiently learn new knowledge, we design a novel paradigm with three methods to step-by-step utilize attention mechanism and prior-knowledge in meta-learning.
	Firstly, the method AML is designed to leverage attention mechanism in meta-learning.
	Secondly, the method RAML is designed to use not only the attention mechanism but also prior-knowledge in meta-learning.
	Compared with RAML, the method URAML learns the prior-knowledge with unsupervised learning, which brings URAML the advantage that with the growth of available unlabeled images used in the prior-knowledge learning stage and the progress of unsupervised learning algorithm, the performance of the meta-learner will be boosted up.

	\section{Experiments}
	In this section, we firstly present the datasets we used in our experiments, and then the details and results of our experiments. 
	
	\subsection{Dataset}
	We use several datasets in all our experiments: MiniImagenet\cite{miniimagenet}, Omniglot\cite{omniglot}, MiniImagenet-900, Places2\cite{places365}, COCO\cite{COCO}, and OpenImages-300.
	Note that, we resize all the images in Omniglot into 28x28 resolution, and all the other images into 84x84.
	
	\subsubsection{MiniImagenet} 
	MiniImagenet\cite{miniimagenet} is popularly used for evaluating few-shot learning and meta-learning. 
	It contains 100 image classes, including 64 training classes, 16 validation classes, and 20 testing classes. 
	Each image class with 600 images are sampled from the ImageNet dataset\cite{imagenet}.
	
	\subsubsection{Omniglot} 
	Omniglot\cite{omniglot} is another widely used dataset for meta-learning. 
	It contains 50 different alphabets and 1623 characters from these alphabets, and each character has 20 images that hand-drawn by 20 different people. 
	
	\subsubsection{MiniImagenet-900} 
	MiniImagenet-900 dataset is designed for the Representation modules in RAML and URAML learning prior-knowledge, and it is composed of 900 image classes. 
	Each image class with 1300 images are collected from the original ImageNet dataset.
	It is worth noting that there is no image class in MiniImageNet-900 coincides with the classes in the MiniImagenet dataset.
	
	\subsubsection{Other Datasets}
	As the Representation module of URAML is trained by unsupervised learning, we take full advantage of this characteristic by training the Representation module of URAML on not only MiniImagenet-900 but also Places2\cite{places365}, COCO2017\cite{COCO}, and OpenImages-300. 
	The dataset OpenImages-300 is a subset of the OpenImages-V4 dataset\cite{openimages}. 
	The total OpenImages-V4 dataset contains 9 million images, and we randomly downloaded 3 million images from the OpenImages-V4 website to form the OpenImages-300 dataset.

	\subsection{Experiments on MiniImagenet}
	On MiniImagenet, we test all our methods on 5-way 1-shot and 5-way 5-shot classification tasks. 
	The testing accuracy is averaged by the accuracies on 600 tasks, with 95\% confidence intervals, and all these 600 tasks are randomly generated on the test set of MiniImagenet. 
	The support and query set of each $N$-way $K$-shot task contains $NK$ and $15*N$ images, respectively.

	\begin{table*}
		\centering
		\caption{Few-shot learning performance on Omniglot. 
			The method which is colored with blue uses deep network (ResNet) to extract image features, while the other use shallow network (4 cascading convolution layers). 
			The accuracy is tested as the same way as MAML\cite{MAML}}
		\resizebox{1.7\columnwidth}{!}{
			\begin{tabular}{|c|c|c|c|c|c|}
				\hline
				\multirow{2}{*}{Method} &\multirow{2}{*}{Venue} &\multicolumn{2}{c|}{5-way Accuracy} &\multicolumn{2}{c|}{20-way Accuracy} \\
				\cline{3-6} & &1-shot &5-shot &1-shot &5-shot\\
				\hline
				MAML\cite{MAML}	&ICML-17	            &98.70$\pm$0.40\%	&99.90$\pm$0.10\%	&95.80$\pm$0.30\%	&98.90$\pm$0.20\% \\
				\hline
				Prototypical Nets\cite{prototypical}	&NIPS-17	&98.80\%	&99.70\%	&96.00\%	  &98.90\% \\
				\hline
				Meta-SGD\cite{Meta-SGD}	&/&99.53$\pm$0.26\%	&{\bfseries 99.93$\pm$0.09\%} &95.93$\pm$0.38\% &98.97$\pm$0.19\% \\
				\hline
				Relation Net\cite{comparenet}	&CVPR-18	    &99.60$\pm$0.20\%	 &99.80$\pm$0.10\%   &97.60$\pm$0.20\%	&99.10$\pm$0.10\%  \\
				\hline
				GNN\cite{GNN}	&ICLR-18	            &99.20\%	&99.70\%	&97.40\%	&99.00\%  \\
				\hline
				Spot-Learn\cite{chu2019spot}	&CVPR-19	            &97.56$\pm$0.31\%	&99.65$\pm$0.06\%	&/	&/  \\
				\hline
				iMAML HF\cite{IMAML}                 &NIPS-19	&{99.50$\pm$0.26\%}	 &99.74$\pm$0.11\%	 &{96.18$\pm$0.36\%}  &{99.14$\pm$0.10\%} \\
				\hline
				{\color{blue}SNAIL}\cite{SNAIL}&ICLR-18 &99.07$\pm$0.16\% &99.78$\pm$0.09\%&97.64$\pm$0.30\% &99.36$\pm$0.18\% \\
				\hline
				{\color{blue}MetaGAN+RN}\cite{Metagan}& NIPS-18 &{\bfseries{\color{black}99.67$\pm$0.18\%}} & 99.86$\pm$0.11\% & 97.64$\pm$0.17\% & 99.21$\pm$0.10\%\\
				\hline
				
				AML(ours) &/	            &{\bfseries 99.65$\pm$0.10\%}	 &99.85$\pm$0.04\%	 &{\bfseries 98.48$\pm$0.09\%}  &{\bfseries 99.55$\pm$0.06\%} \\
				\hline
			\end{tabular}
		}
		\label{tab:result on Omniglot}
	\end{table*}

	\begin{table}
		\centering
		\caption{Few-shot learning performance on MiniImagenet. 
			The method which is colored with blue uses deep network to extract image features, while the other use shallow network. 
			We separately highlight the best result of the methods using shallow network and that of the methods using deep network, for each task. 
		}
		\resizebox{0.98\columnwidth}{!}{
			\begin{tabular}{|c|c|c|c|}
				\hline
				\multirow{2}{*}{Method} &\multirow{2}{*}{Venue} &\multicolumn{2}{c|}{5-way Accuracy} \\
				\cline{3-4} & &1-shot &5-shot \\
				\hline
				MAML\cite{MAML} &ICML-17	             &48.70$\pm$1.84\% &63.11$\pm$0.92\% \\
				\hline
				Prototypical Nets\cite{prototypical}	&NIPS-17 &49.42$\pm$0.78\%	&68.20$\pm$0.66\% \\
				\hline
				Meta-SGD\cite{Meta-SGD} &        /         &50.47$\pm$1.87\%	&64.03$\pm$0.94\% \\
				\hline
				LLAMA\cite{LLAML}          & ICLR-18   & 49.40$\pm$1.83\% & / \\
				\hline
				Relation Net\cite{comparenet} &CVPR-18	     &51.38$\pm$0.82\%  &67.07$\pm$0.69\% \\
				\hline
				GNN\cite{GNN} &ICLR-18	             &50.33$\pm$0.36\%	&66.41$\pm$0.63\% \\
				\hline
				Spot-Learn\cite{chu2019spot} &CVPR-19	             &51.03$\pm$0.78\%	&67.96$\pm$0.71\% \\
				\hline
				iMAML HF\cite{IMAML}  &NIPS-19 &49.30$\pm$1.88\%	&/ \\
				\hline
				Meta-MinibatchProx\cite{zhou2019efficient}  &NIPS-19 &50.77$\pm$0.90\%	&67.43$\pm$0.89 \\
				\hline
				AML(ours)&	/	&{\bfseries {\color{black} 52.25$\pm$0.85\%}}	&{\bfseries {\color{black} 69.46$\pm$0.68\%} } \\
				\hline
				\hline
				{\color{blue}SNAIL}\cite{SNAIL}& ICLR-18 &55.71$\pm$0.99\%  &68.88$\pm$0.92\% \\
				\hline
				{\color{blue}TADAM}\cite{TADAM}& NIPS-18 &58.50$\pm$0.30\%  &76.70$\pm$0.30\% \\
				\hline
				{\color{blue}MetaGAN+RN}\cite{Metagan}& NIPS-18 &52.71$\pm$0.64\% & 68.63$\pm$0.67\%  \\
				\hline
				{\color{blue}AM3-TADAM}\cite{xing2019adaptive}& ICLR-19 &65.30$\pm$0.49\%  &78.10$\pm$0.36\% \\
				\hline
				{\color{blue}Incremental}\cite{ren2019incremental}& NIPS-19 &54.95$\pm$0.30\%  &63.04$\pm$0.30\% \\
				\hline
				{\color{blue}RAML(ours)}	&/&{\bfseries {\color{black}63.66$\pm$0.85\%}}	&{\bfseries {\color{black}80.49$\pm$0.45\%}} \\
				\hline
				{\color{blue}URAML(ours)}	&/&{\bfseries {\color{black}49.56$\pm$0.79\%}}	&{\bfseries {\color{black}63.42$\pm$0.76\%}} \\
				\hline
			\end{tabular}
		}
		\label{tab:result on MiniImagenet}
	\end{table}

	\textbf{In AML}, the network structure of the meta-leaner is shown in Fig.\ref{fig:network of the AML method}. 
	The feature extractor is composed of 4 Convolution layers and the classifier is a fully-connect layer, and the attention model structure is shown in Fig.\ref{fig:network structure of attention}.
	Each Convolution layer consists of 64 channels and is followed with a ReLU and batch-normalization layer.
	We train the meta-learner on 200000 randomly generated tasks for 60000 iterations, and set the learning rate to 0.001, and decay the learning rate to 0.0001 after 30000 iterations. 
	Moreover, Dropout with dropout-rate 0.2, L1 and L2 normalization with 0.001 and 0.00001, respectively, are used to prevent the meta-learner from over-fitting.
	
	The experimental result of the method AML on MiniImagenet shows in Tab.\ref{tab:result on MiniImagenet}.
	Note that in Tab.\ref{tab:result on MiniImagenet}, the method whose name is printed as black uses a shallow network consists of 4 or 5 Convolution layers and one or two fully-connect layers, and the method whose name is printed as blue uses a deep ResNet-based network.
	Among all the methods using shallow network, AML attained the state-of-the-art on both the 5-way 1-shot and 5-way 5-shot image classification tasks.
	
	\textbf{In RAML}, the Representation module is a ResNet-50\cite{resnet} network, and the Auxiliary module is a fully-connect layer. 
	The attention model is the same as that in AML, and the classifier is composed of two fully-connect layers. 
	
	At the prior-knowledge learning stage, we set the batch size to 256, and the learning rate to 0.001, and decay the learning rate to 0.0001 after 30000 iterations, and use L2 normalization with 0.00001 and Dropout with 0.2 to prevent the Representation module from over-fitting.
	At the meta-training stage, the ABP module is meta-trained with the same setting as AML. 
	The experiment result of RAML is shown in Tab.\ref{tab:result on MiniImagenet}. 
	Compared to method AML, RAML improves the meta-learner's performance more significantly.
	It rises the accuracy on 5-way 1-shot tasks from 52.25\% to 63.66\%, and the accuracy on 5-way 5-shot tasks from 69.46\% to 80.49\%.
	
	The most likely reason why RAML performs well is: before the meta-training stage, the Representation module has learned old knowledge to help the meta-learner understanding new input image and provides high-level meaningful representations and features of the input image.
	In the meta-training stage, the meta-learner's work becomes more comfortable because it only needs to learn how to quickly adjust its ABP module according to the compact features the Representation module provided, and do not need to take care of the original high dimensional input data. 
	While the meta-learner of AML works harder than the meta-learner of RAML, as it has to adjust its total network to fit new few-shot learning tasks according to the original input data.

	\begin{table*}
		\centering
		\caption{Ablation experimental results about the attention mechanism on Omniglot.}
		\resizebox{1.4\columnwidth}{!}{
			\begin{tabular}{|l|c|c|c|c|}
				\hline
				\multirow{2}{*}{Method} &\multicolumn{2}{c|}{5-way Accuracy} &\multicolumn{2}{c|}{20-way Accuracy} \\
				\cline{2-5} &1-shot &5-shot &1-shot &5-shot \\
				\hline
				MAML* &97.40$\pm$0.27\% &{\bfseries 99.71$\pm$0.05\%} &{\bfseries 93.37$\pm$0.23\%} &97.46$\pm$0.11\% \\
				\hline
				MAML+attention &{\bfseries 97.41$\pm$0.28\%} &99.48$\pm$0.12\% &92.99$\pm$0.25\% &{\bfseries 97.94$\pm$0.10\%} \\
				\hline
				Meta-SGD* &98.94$\pm$0.17\% &99.51$\pm$0.07\% &95.82$\pm$0.21\% &98.40$\pm$0.09\% \\
				\hline
				Meta-SGD+attention &{\bfseries 99.26$\pm$0.15\%} &{\bfseries 99.79$\pm$0.04\%} &{\bfseries 97.94$\pm$0.14\%} &{\bfseries 98.99$\pm$0.10\%} \\
				\hline
			\end{tabular}
		}
		\label{tab:ablation result on Omniglot}
	\end{table*}
	
	\textbf{In URAML}, the Representation module learns the prior-knowledge with an unsupervised learning algorithm: Split-Brain. 
	As Fig.\ref{fig:figure5} shows, two independent ResNet-50 network-based encoders compose the Representation module, and we halve all the filters in each encoder so that the Representation module outputs feature vector with a dimension of 2048, which is the same with that in RAML.
	The Auxiliary module is composed of two deconvolution-based decoders, and Tab.\ref{tab:URAML_decoder_net} shows the detail of the decoder network structure. 
	The last Conv-layer's number of filters is 1 or 2 according to that the decoder is recovering the \emph{L} channel or the \emph{ab} channels of the \emph{Lab} image.
	
	At both the prior-knowledge learning and meta-training stage, we set all hyperparameters the same with those in the RAML experiment. 
	Noted that for saving the training computation cost, the decoders in the Auxiliary module recover the \emph{ab} and \emph{L} channels into 11x11 resolution, but not the original 84x84.
	When calculating the \emph{MSE} losses $L_{l}(\theta_l, \omega_{l})$ and $L_{ab}(\theta_{ab}, \omega_{ab})$ shown in Eq.\ref{eq:URAML_knowledge1} and Eq.\ref{eq:URAML_knowledge2}, we first resize \emph{ab} and \emph{L} channels of the input \emph{Lab} image into 11x11 resolution and then calculate $L_{l}(\theta_l, \omega_{l})$ and $L_{ab}(\theta_{ab}, \omega_{ab})$. 
	The experiment result of URAML is shown in Tab.\ref{tab:result on MiniImagenet}. 
	We also highlight the result of URAML in Tab.\ref{tab:result on MiniImagenet}, even though its result is not state-of-the-art.
	In our viewpoint, the reason why URAML lags behind RAML is that the Representation module in URAML learns the prior-knowledge with unsupervised learning while the Representation module in RAML learns with supervised learning.
	
	\begin{table}[!t]
		\renewcommand{\arraystretch}{1.3}
		\caption{Detailed structure of the decoder module in URAML.}
		\centering
		\label{tab:URAML_decoder_net}
		\resizebox{0.6\columnwidth}{!}{
			\begin{tabular}{|c|c|c|c|}
				\hline
				Layers & Number of filters & Kernel \\
				\hline
				CONV & 1024 &5 \\
				\hline
				DeCONV & 512 &3 \\
				\hline
				DeCONV & 256 &3 \\
				\hline
				CONV & 1 or 2 &1 \\
				\hline
			\end{tabular}
		}
	\end{table}

	\begin{table}
		\centering
		\caption{Ablation experimental results about the attention mechanism on MiniImagenet}
		\resizebox{0.9\columnwidth}{!}{
			\begin{tabular}{|l|c|c|}
				\hline
				\multirow{2}{*}{Method} &\multicolumn{2}{c|}{5-way Accuracy} \\
				\cline{2-3} &1-shot &5-shot \\
				\hline
				MAML* &48.03$\pm$0.83\% &64.11$\pm$0.73\% \\
				\hline
				MAML+attention &{\bfseries 48.52$\pm$0.85\%} &{\bfseries 64.94$\pm$0.69\%} \\
				\hline
				Reptile* &48.23$\pm$0.43\% &63.69$\pm$0.49\% \\
				\hline
				Reptile+attention &{\bfseries 48.30$\pm$0.45\%} &{\bfseries 64.22$\pm$0.39\%} \\
				\hline
				Meta-SGD* &48.15$\pm$0.93\% &63.73$\pm$0.85\% \\
				\hline
				Meta-SGD+attention &{\bfseries 49.11$\pm$0.94\%} &{\bfseries 65.54$\pm$0.84\%} \\
				\hline
			\end{tabular}
		}
		\label{tab:ablation result on MiniImagenet}
	\end{table}

	\subsection{Experiments on Omniglot}
	As Omniglot is a much easier dataset than MiniImagenet that existing meta-learners can easily achieve more than 95\% accuracy on most testing tasks generated on Omniglot, we only test method AML on Omniglot. 
	
	Same to the experiments on Miniimagenet, we also train the meta-learner on 200000 randomly generated tasks for 60000 iterations and set the learning rate to 0.001.
	The experiment results are shown in Tab.\ref{tab:result on Omniglot}
	
	It is clear that the proposed method AML attains state-of-the-art performance on 2 of all 4 kinds of few-shot image classification tasks. 
	On the 5-way 1-shot task, though the method MetaGAN+RN performs slightly better than AML, we still highlight AML as MetaGAN+RN uses a deeper ResNet-based network while AML uses a shallower network.
	On the 20-way 1-shot task, our method AML surpasses other methods by a large margin.
	For example, compared to IMAML HF, AML improves the meta-learner's performance from 96.18\% to 98.48\%.

	\subsection{Ablation Study}
	\subsubsection{Ablation Study about the Attention Mechanism}
	To confirm the promotion effect of the attention mechanism for meta-learning, we conduct experiments to compare the performance of the meta-learner which is equipped with the attention model and its counterpart which is not. 
	The experimental results show in Tab.\ref{tab:ablation result on MiniImagenet} and Tab.\ref{tab:ablation result on Omniglot}. 
	The compared meta-learner which is marked with * is the meta-learner re-implemented by ourselves. 
	The performances of our re-implemented meta-learners differ slightly from those reported in their original papers.
	This is probably caused by different hyper-parameters or experiment settings (all methods in this experiment use convolution layers with 32 filters). 
	The comparisons in Tab.\ref{tab:ablation result on MiniImagenet} and Tab.\ref{tab:ablation result on Omniglot} revealing that in most cases, the attention mechanism improves the meta-learner significantly, which demonstrates the reason-ability of our idea. 
	
	As attention mechanism brings the meta-learner more weights and computation cost, we do another experiment to validate that the improvement of AML is the contribution of the attention mechanism but not the growth of the number of weights and computation cost. 
	The experiment detail is: since the attention model in AML is a convolution layer with the kernel size of 1x1, we remove the attention model, and stack a convolution layer with the same kernel size on the top of the CNN feature extractor.
	We name the meta-learner with this network as AML-attention, and its number of weight is the same as that of AML. 
	The corresponding experimental result is shown in Tab.\ref{tab:other ablation}, and it is clear that AML outperforms AML-attention, which further shows the improvement effect of attention mechanism for meta-learning.

	\begin{table}
		\centering
		\caption{Results of several ablation experiments. 
		}
		\resizebox{0.8\columnwidth}{!}{
			\begin{tabular}{|l|c|c|}
				\hline
				\multirow{2}{*}{Method} &\multicolumn{2}{c|}{5-way Accuracy} \\
				\cline{2-3} &1-shot &5-shot \\
				\hline
				AML&{\bfseries 52.25$\pm$0.85\%}	&{\bfseries69.46$\pm$0.68\%} \\
				\hline
				AML-attention &51.27$\pm$0.78\%	&67.73$\pm$0.65\% \\
				\hline
				RAML	&{\bfseries 63.66$\pm$0.85\%}	& {\bfseries 80.49$\pm$0.45\%} \\
				\hline
				RAML-Places2	&58.82$\pm$0.89\%	&74.09$\pm$0.76\% \\
				\hline
			\end{tabular}
		}
		
		\label{tab:other ablation}
	\end{table}
	
	\begin{table*}
		\centering
		\footnotesize
		\caption{Ablation experimental results about URAML.}
		\resizebox{1.9\columnwidth}{!}{
			\begin{tabular}{|l|c|c|c|c|}
				\hline
				\multirow{2}{*}{Method} & \multirow{2}{*}{Dataset} & \multirow{2}{*}{Number of images} & \multicolumn{2}{c|}{5-way Accuracy} \\
				\cline{4-5} & & &1-shot &5-shot \\
				\hline
				URAML-V1 & MiniImagenet-900  & 1.15million	&45.91$\pm$0.79\%   &61.04$\pm$0.71\% \\
				\hline
				URAML-V2 & MiniImagenet-900, places365, COCO2017 & 4.10million	&48.82$\pm$0.79\%	&62.84$\pm$0.78\% \\
				\hline
				URAML-AE & MiniImagenet-900, places365, COCO2017, OpenImages-300 & 7.10million	&33.29$\pm$0.71\%   &43.60$\pm$0.66\% \\
				\hline
				URAML & MiniImagenet-900, places365, COCO2017, OpenImages-300 & 7.10million	&{\bfseries 49.56$\pm$0.79\%}	&{\bfseries63.42$\pm$0.76\%} \\
				\hline
			\end{tabular}
		}
		
		\label{tab:URAML_exp}
	\end{table*}
	
	\subsubsection{Prior-Knowledge Learning Dataset}
	We do experiments to test how does the prior-knowledge learning dataset affects RAML and URAML. 
	
	\emph{a) affects to RAML}: 
	In RAML, the default prior-knowledge learning dataset is our reorganized \emph{Miniimagenet-900} dataset.
	In this experiment, the Representation module learns the prior-knowledge on Places2\cite{places365} instead of \emph{Miniimagenet-900}, and all the other experiment settings and hyper-parameters are constant with the primordial RAML.
	We denote this meta-learner as RAML-Places2. 
	Corresponding experimental result shows in Tab.\ref{tab:other ablation}.
	It is clear that prior-knowledge learning dataset affects the meta-learner. 
	The reason is that different prior-knowledge learning dataset leads the Representation module learning different knowledge and expressing image features differently.
	Places2 is a dataset commonly used for scene classification, which results in that the Representation module learning the knowledge about scene understanding rather than object classification. 
	
	\emph{b) affects to URAML}: 
	In this experiment, we test how the quantity of unlabeled \emph{Lab} images in the prior-knowledge learning dataset affect URAML. 
	We design two new versions of URAML: URAML-V1 and URAML-V2. 
	The Representation module of URAML-V1 learns prior-knowledge only on MiniImagenet-900, and that of URAML-V2 learns prior-knowledge on not only MiniImagenet-900, but also the Places2 and COCO2017. 
	Compared with URAML-V1 and URAML-V2, the quantity of unlabeled \emph{Lab} used in the primordial URAML is the largest, as MiniImagenet-900, places365, COCO2017, and OpenImages-300 are all used in the primordial URAML. 
	Tab.\ref{tab:URAML_exp} shows the prior-knowledge learning dataset and the performances of URAML-V1, URAML-V2, and the primordial URAML.
	It is clear that the primordial URAML performs the best, and the more the unlabeled \emph{Lab} images used for the meta-learner to learn prior-knowledge, the better the meta-learner performs.
	Besides, there remains a large performance progress space as we can use more unlabeled data in URAML.

	\subsubsection{Unsupervised Learning for URAML}
	The development of unsupervised learning also affects URAML a lot. 
	To verify this viewpoint, we do an experiment that the Representation module in URAML learns the prior-knowledge with a basic unsupervised learning method Auto-Encoder\cite{AE}, and we name this version of URAML as URAML-AE. 
	The experimental result of URAML-AE shown in Tab.\ref{tab:URAML_exp} revealing that the unsupervised learning algorithm affects the meta-learner significantly.
	Maybe the most promising way to improve the performance of URAML is to develop the unsupervised learning algorithm and collect more unlabeled data.

	\subsection{Cross-Testing Experiment}
	
	We find that existing meta-learning methods generally suffer from a Task-Over-Fitting (TOF) problem, and this problem has seldom been studied.
	An example of the TOF problem is that the meta-learner to be tested on 5-way 1-shot classification tasks should be trained on 5-way 1-shot tasks rather than on other tasks, and similarly, the meta-learner to be tested on 5-way 5-shot tasks should be trained on 5-way 5-shot tasks. 
	This is because the meta-learner trained on 5-shot tasks over-fits to 5-shot tasks, and when testing it on 1-shot tasks, it will perform obviously worse than the meta-learner trained on 1-shot tasks.
	
	We do lots of cross-testing experiments to test how much does MAML, Meta-SGD, AML, RAML, and URAML suffer from the TOF problem, and the experimental results show that compared with the other methods, our methods suffer less from this problem, especially RAML and URAML.
	
\begin{figure*}
		\centering
		\includegraphics[width=2.05\columnwidth]{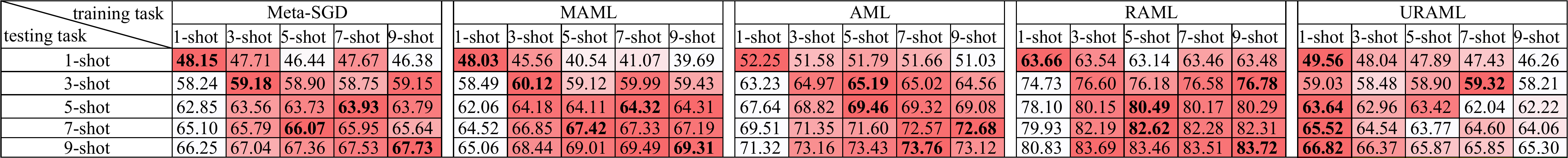}
		\caption{Results of cross-testing experiments amoung MAML, Meta-SGD, AML, RAML and URAML. 
			Each column presents a meta-learner trained on specific $K$-shot training tasks and each row presents specific $J$-shot testing tasks, where $K, J\in$\{1,3,5,7,9\}.
			For each method, the value at $J$-shot row $K$-shot column presents the $J$-shot testing accuracy of the meta-learner trained on $K$-shot training tasks.
			For example, the value 59.99 at 3-shot row 7-shot column of MAML presents the 3-shot testing accuracy of the MAML meta-learner trained on 7-shot training tasks.
			The value 80.83 at 9-shot row 1-shot column of RAML presents the 9-shot testing accuracy of the RAML meta-learner trained on 1-shot training tasks.
		}
		\label{fig:cross_test}
	\end{figure*}

	For each tested meta-learning method, we do the cross-testing experiments in the following way:
	1) train the meta-learner on 5-way $K$-shot image classification tasks, where $K\in$\{1,3,5,7,9\},
	2) test the meta-learner on 5-way $J$-shot tasks, where $J\in$\{1,3,5,7,9\}. 
	For example, we train a meta-learner with MAML on 5-way 3-shot tasks and test its performance on all 5-way $K$-shot tasks, $K\in$\{1,3,5,7,9\}. 
	The experimental results are shown in Fig.\ref{fig:cross_test}.
	
	Obviously, Fig.\ref{fig:cross_test} shows that MAML suffers seriously from the TOF problem, because its meta-learner which performs best on $K$-shot tasks probably performs not well on $J$-shot tasks, where $K \neq J$. 
	For example, in MAML, the meta-learner trained on 1-shot tasks performs best on the 1-shot tasks, but it can not perform as well as the other meta-learners on 3-, 5-, 7-, and 9-shot tasks, which means the meta-learner trained on 1-shot tasks over-fits to 1-shot tasks.
	The meta-learner trained by URAML troubled little by the TOF problem because the meta-learner which performs best on $K$-shot tasks probably performs best on $J$-shot tasks, where $K$, $J\in$\{1,5,7,9\}.
	For example, in URAML, the meta-learner trained on 1-shot tasks performs best not only on the 1-shot tasks but also on 5-, 7-, and 9-shot tasks, which means the meta-learner trained on 1-shot tasks generalizes well to the other $J$-shot tasks.
	
	We design a metric Cross-Entropy across Tasks (CET), to quantize how much does a meta-learning approach be vulnerable to the TOF problem. 
	The evaluation process is shown as Eq.\ref{eq:entropy}, where \emph{i}, \emph{j}$\in$\{1,3,5,7,9\} and overstriking variables are vector.
	$\mathcal{S}$ and $\mathcal{D}$ are the softmax and cross-entropy operation.
	$\bm{a}_i$ is the testing accuracies of five meta-learners trained on 1-, 3-, 5-, 7-, 9-shot tasks when they are tested on $i$-shot tasks.
	$\bm{d}_i$ is the meta-learners' accuracy distribution on $i$-shot tasks.
	$l_{i,j}$ presents the similarity between accuracy distribution vector \emph{\textbf{d$_{i}$}} and \emph{\textbf{d$_{j}$}}, where \emph{i,j$\in${\{1,3,5,7,9\}}}. 
	\emph{L} presents the overall similarities of $l_{i,j}$ for a specific approach. 
	
	\begin{equation}
	\left\{
	\begin{array}{lr}
	\bm{d}_i = \mathcal{S}(\bm{a_i}/max(\bm{a_i})) \\
	l_{ij} = \mathcal{D}(\bm{d_i}, \bm{d_j}) \\
	L = \sum_{i,j\in{1,3,5,7,9}}^{i\neq j}l_{ij}
	\end{array}
	\right.
	\label{eq:entropy}
	\end{equation}
	
	For example, the testing accuracies $\bm{a}_{3}$ of Meta-SGD [58.24\%, 59.18\%, 58.90\%, 58.75\%, 59.15\%] is the five trained meta-learners of Meta-SGD when they are tested on 3-shot tasks. 
	So, $\bm{a}_3 / max(\bm{a}_3)$ = [58.24\%, 59.18\%, 58.90\%, 58.75\%, 59.15\%] / 59.18\%, and $\bm{d}_{3}$ = $\mathcal{S}(\bm{a}_3 / max(\bm{a}_3))$ = [0.116, 0.255, 0.202, 0.178, 0.249].
	Similarly, \emph{\textbf{d$_{7}$}} = [0.122, 0.206, 0.255, 0.233, 0.184].
	Then, \emph{l$_{3,7}$} = 1.603, and \emph{L} = 34.22.
	
	Obviously, the smaller the total distance \emph{L} appears, the less the meta-learning approach suffers from the TOF problem.
	We show different meta-learning approaches' performance on the CET metric in Tab.\ref{tab:cross testing}. 
	This experiment shows that the proposed AML, RAML, and URAML performs better then MAML and Meta-SGD on the CET metric, and RAML and URAML performs best.
	The possible reason for this is that prior-knowledge and attention mechanism are both helpful for the meta-learner to reduce its few-shot cognitive load and to avoid itself be affected by redundant useless information.

	\begin{table}
		\centering
		\caption{Performance of different meta-learning methods on the CET metric.}
		\resizebox{0.9\columnwidth}{!}{
			\begin{tabular}{|c|c|c|c|c|c|}
				\hline
				Method	& MAML & Meta-SGD & AML & RAML &URAML \\
				\hline
				CET & 57.19 & 34.22 & 33.35 & {\bfseries 32.13} &  32.16 \\ 
				\hline
			\end{tabular}
		}
		\label{tab:cross testing}
	\end{table}

	We can see an interesting phenomenon in Fig.\ref{fig:cross_test}, that the meta-learner trained by RAML on 5-way 9shot tasks performs best in most of the test tasks, while the meta-learner trained by URAML on 5-way 1-shot tasks performs best. 
	The possible reason behind this phenomenon is that the Representation module of RAML learns knowledge by supervised learning, while the Representation module of URAML learns knowledge by unsupervised learning, which results in the output features between these two kinds of Representation module be different.

	\subsection{Feature Analysis}
\begin{figure*}
		\centering
		\includegraphics[width=2.05\columnwidth]{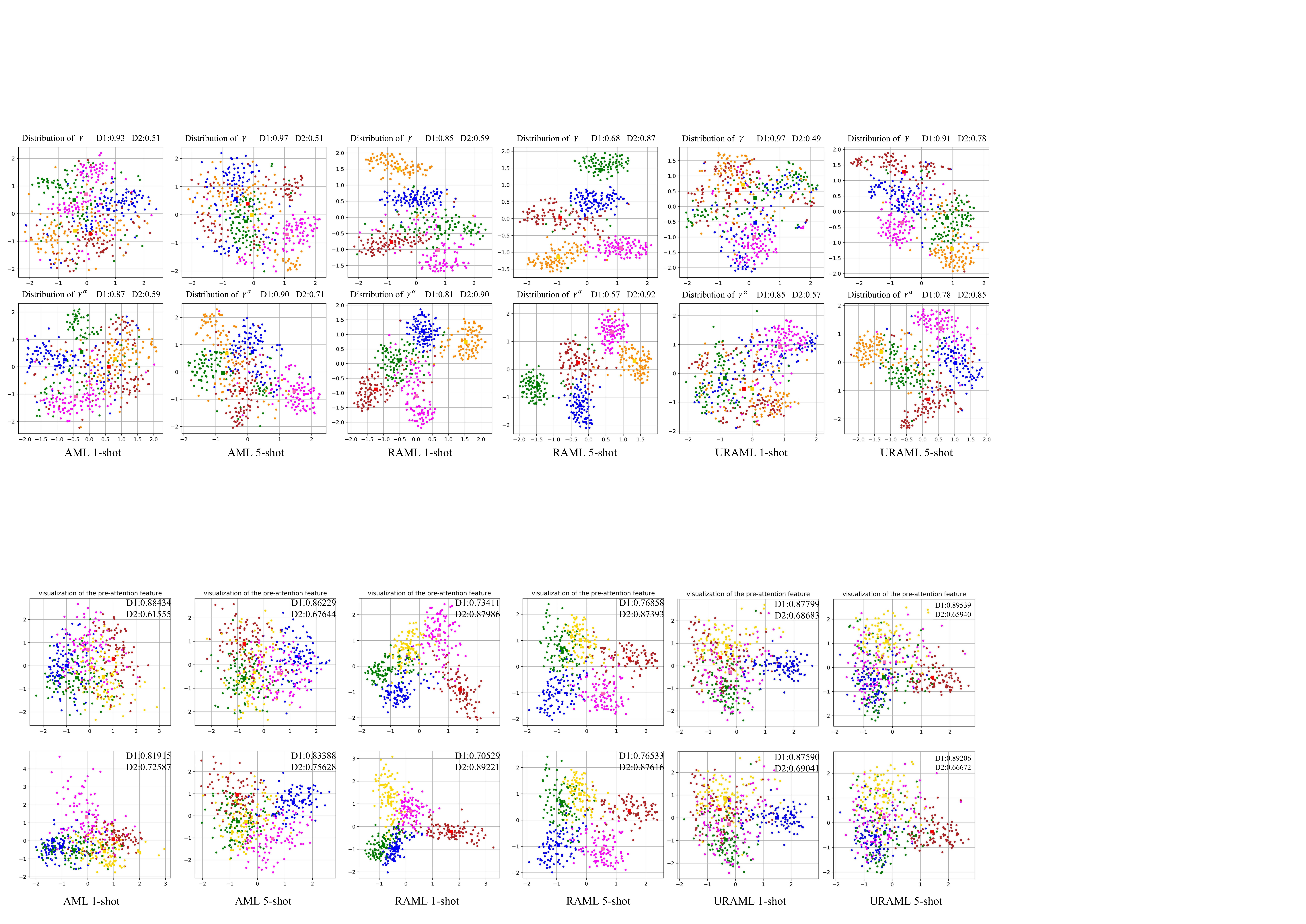}
		\caption{
			Visualization of the distributions of features $\gamma$ and $\gamma^\alpha$ of all our three methods. 
			For each method, we randomly generate a 5-way 1-shot and a 5-way 5-shot testing task on Miniimagenet, and the query set of each task contains 100 images for each image class.
			For each testing task, after the meta-learner inner-updating on the support set, we use t-SNE to visualize the distributions of the meta-learner's $\gamma$ and $\gamma^\alpha$ of the query set images.
			For each picture, five colors are used to represents 5 image classes in the testing task and each point denotes the feature $\gamma$ or $\gamma^\alpha$.
			We also show the average distribution inner-class distance D1 and inter-class distance D2 above each picture to better understand the distributions. 
		}
		\label{fig:feature points}
	\end{figure*}

	To understand the effect of attention mechanism, we visualize the distributions of feature $\gamma$ and $\gamma^\alpha$ (shown in Fig.\ref{fig:network of the AML method}, Fig.\ref{fig:4a} and Fig.\ref{fig:figure5}) in Fig.\ref{fig:feature points} with t-SNE\cite{tSNE}. 
	In Fig.\ref{fig:feature points}, 500 feature points of each picture represent 500 $\gamma$ or $\gamma^\alpha$ of the query set images of a 5-way 1 or 5 shot task that randomly generated on the test set of MiniImagenet.
	
	The average distribution inner-class distance D1 of $\gamma^\alpha$ is smaller than that of $\gamma$, and the average inter-class distance D2 of $\gamma^\alpha$ is larger than that of $\gamma$.
	This result indicates that among different image classes, the distribution of $\gamma^\alpha$ is more distinguishable than that of $\gamma$.
	The reason for this is that the attention mechanism makes the meta-learner be able to adjust its attention quickly to critical image features and makes $\gamma^\alpha$ more distinguishable than $\gamma$ to differentiate images of different classes.

	\subsection{Heat-Map of $\gamma$ and $\gamma^\alpha$}
\begin{figure}
		\centering
		\includegraphics[width=1\columnwidth]{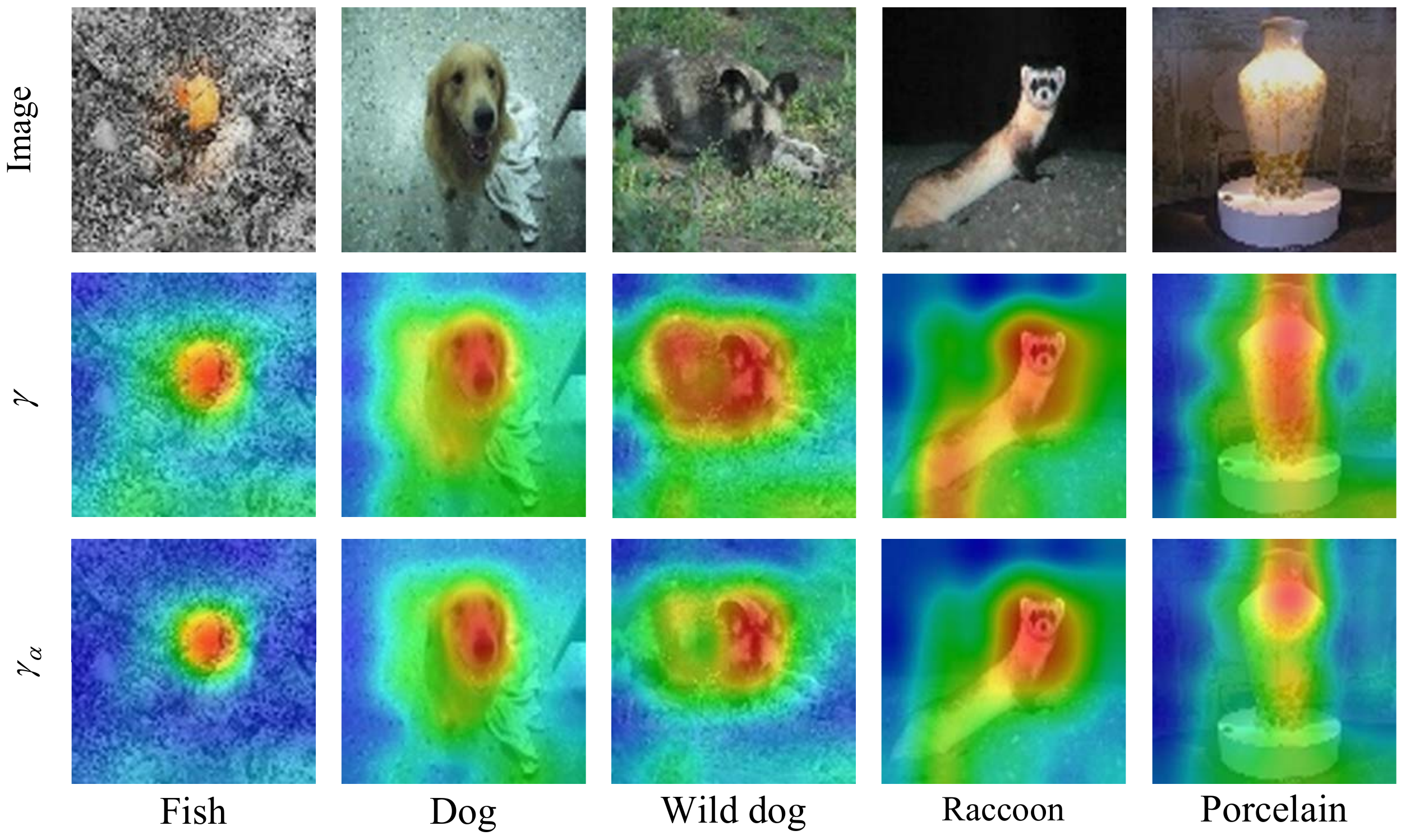}
		\caption{We show some images which are sampled from the query set of a 5-way 1-shot classification task, and the corresponding heat-maps of $\gamma$ and $\gamma^\alpha$.}
		\label{fig:att_heat map}
	\end{figure}

	To further analyze how the attention mechanism affects the meta-learner, we visualize the heat-maps of $\gamma$ and $\gamma^\alpha$ in Fig.\ref{fig:att_heat map}. 
	To get the heat-map of $\gamma$, we first inner-update the RAML meta-learner on the support set of a randomly generated 5-way 1-shot testing task on MiniImagenet.
	Then, we feed the meta-learner with the query set images and average the feature maps $\gamma$ across the channel axis to get the heat-maps of $\gamma$.
	Similarly, the heat-maps of $\gamma^\alpha$ can be got.
	
	From the heat-maps shown in Fig.\ref{fig:att_heat map}, we can see that compared with $\gamma$, $\gamma^\alpha$ is more sensitive to the distinguishable part of the input image, revealing that the meta-learner changes its attention to the most discriminative image feature. 
	For example, the first column of Fig.\ref{fig:att_heat map} is a fish. 
	Besides the fish body, $\gamma$ is also sensitive to some background region of the image. 
	However, the meta-learner discovers that only the fish body is the crucial feature to category this image and shrinks its attention region so that $\gamma^\alpha$ sensitive only to the fish body.
	
	Through the visualization and analysis of the heat-map of $\gamma$ and $\gamma^\alpha$, we can see that the attention mechanism helps the meta-learner to focus on the most distinguishable image feature, and further helps the meta-learner to do a better few-shot learning task.
	
	\section{Conclusion and the Future Work}
	In this paper, be inspired by human cognition and learning process, we find the importance of attention mechanism and the prior-knowledge for meta-learning based few-shot learning.
	To solve a few-shot learning task, the meta-learner should first well use stable prior-knowledge to understand images and extract compact feature representations of images so that it can solve the task in the compact representation space rather than the original image space.
	Then, the meta-learner should adjust its attention to the crucial feature of the extracted feature representations, and make the final decision based on its attention.
	Therefore, we step-by-step propose three methods AML, RAML, and URAML to introduce attention mechanism and the prior-knowledge to meta-learning. 
	All of them work successfully with state-of-the-art performance on several few-shot learning benchmarks, which indicating the rationality of our viewpoints and methods. 
	
	Besides, we find existing meta-learning approaches suffer from the TOF problem, which is unfriendly to practical applications. 
	We design a novel Cross-Entropy across Tasks (CET) metric to evaluate how much does a meta-learning suffers from TOF.
	The experiment shows that compared to existing meta-learning methods, the proposed methods suffer less from the TOF problem, especially the RAML and URAML methods. 
	
	Among all the proposed methods, though URAML performs not the best, we think it is the most promising method yet because there is a large development space for the performance of URAML method which will also be the direction of our future work. 
	From the ablation study, two manners seem can improve the performance of URAML significantly. 
	One is to develop the unsupervised learning algorithm or self-supervised learning. 
	RAML performs better than URAML revealing that the current unsupervised learning algorithm falls behind supervised learning. 
	Bridging the gap between unsupervised learning and supervised learning algorithms will boost up the performance of URAML in a substantial probability. 
	The other manner is to use more unlabeled data for URAML to learn prior-knowledge. 
	Although 7.1 million unlabeled images are used in URAML, it still dramatically falls behind the images that humans have ever seen in terms of both quantity and quality. 
	As for the quantity, we assume that, if a person watches 1 image per second and keep watching 15 hours per day, he/she can see 100 million images in 5 years.
	As for quality, humans see the world in a multimodal way, that is, the human can not only see the object but also touch and move around the object, which helps humans understand the world more accurately than Computer Vision.
	In a word, developing the unsupervised or self-supervised learning algorithm and collecting more unlabeled images will both help URAML to perform well.

	\section*{Acknowledgements}
	This work is supported by the National Natural Science Foundation of China (No. 61573286).
	


	\bibliographystyle{IEEEtran}
	
	\bibliography{IEEEabrv,main}
	%


	
	

\end{document}